\documentclass[journal]{IEEEtran}
\usepackage{xcolor}
\usepackage{setspace}
\usepackage{float}
\usepackage[fleqn]{amsmath}
\usepackage{enumerate}%\begin{enumerate}[i] % reacts to 1, a, A, i, and I
\usepackage{mathrsfs}
\usepackage{makeidx}         % allows index generation
\usepackage{graphicx}        % standard LaTeX graphics tool
\usepackage{multicol}        % used for the two-column index
\usepackage{subfigure}
\usepackage{epsfig,amssymb,latexsym}
\usepackage{psfrag}
\usepackage[ruled]{algorithm}
\usepackage{algorithmic}
\usepackage{url}
\newcommand{\be}{\begin{equation}}
\newcommand{\ee}{\end{equation}}
\newcommand{\bea}{\begin{eqnarray}}
\newcommand{\eea}{\end{eqnarray}}
\newcommand{\bal}{\begin{align}}
\newcommand{\eal}{\end{align}}
\newcommand{\ba}{\begin{array}}
\newcommand{\ea}{\end{array}}
\newcommand{\bc}{\begin{center}}
\newcommand{\ec}{\end{center}}

\allowdisplaybreaks

\ifCLASSINFOpdf
\else
\fi

\begin{document}
\title{Shallow Cue Guided Deep Visual Tracking via Mixed Models}

\author{Fangwen~Tu,
       Shuzhi~Sam~Ge,~\IEEEmembership{Fellow,~IEEE,}
        and~Chang~Chieh~Hang,~\IEEEmembership{Fellow,~IEEE}% <-this % stops a space

\thanks{F. Tu is with the Department of Electrical and Computer Engineering, National
University of Singapore, Singapore 117576 (e-mail: $\text{fangwen\_tu@hotmail.com}$).}% <-this % stops a space

\thanks{S. S. Ge is with the Department of Electrical and Computer Engineering, National
University of Singapore, Singapore 117576 and also with the Social Robotics Lab, Interactive Digital
Media Institute (IDMI), National University of Singapore, Singapore 117576, (e-mail: samge@nus.edu.sg).}% <-this % stops a space
}
\maketitle

\begin{abstract}
In this paper, a robust visual tracking approach via mixed model based convolutional neural networks (SDT) is developed. In order to handle abrupt or fast motion, a prior map is generated to facilitate the localization of region of interest (ROI) before the deep tracker is performed. A top-down saliency model with nineteen shallow cues are employed to construct the prior map with online learnt combination weights. Moreover, apart from a holistic deep learner, four local networks are also trained to learn different components of the target. The generated four local heat maps will facilitate to rectify the holistic map by eliminating the distracters to avoid drifting. Furthermore, to guarantee the instance for online update of high quality, a prioritised update strategy is implemented by casting the problem into a label noise problem. The selection probability is designed by considering both confidence values and bio-inspired memory for temporal information integration. Experiments are conducted qualitatively and quantitatively on a set of challenging image sequences. Comparative study demonstrates that the proposed algorithm outperforms other state-of-the-art methods.\\
\end{abstract}

%\begin{IEEEkeywords}
%Shallow cue, prior map, mixed model, CNN architecture, prioritised update, bio-inspired memory, visual tracking
%\end{IEEEkeywords}

% \fi
\IEEEpeerreviewmaketitle

\section{Introduction}
Visual tracking which focuses on estimating the location of designated target in video clips is one of the most important research topic in computer vision. It plays a crucial role in numerous applications such as security surveillance, robotics, motion recognition and analysis, military patrol and etc.\\
\indent Most of the existing trackers depending on either generative model \cite{bao2012real} which performs template matching or discriminative model \cite{zhong2014robust} \cite{hare2016struck} which separates foreground object from background by treating the tracking task as binary classification problem employ low-level hand-crafted features for the model construction. This kind of features are not sufficient to capture the semantic information and less robust to the appearance variation due to the limited discriminative power. Recently, there is an increasing trend to incorporate deep neural networks into visual tracking by exploiting the rich hierarchical features \cite{zhang2016robust}. The superior performance is mainly attributed to the captured sophisticated hierarchies and the capability of semantic information expression. While, the direct utilization of deep neural networks such as CNN for online visual tracking confront several challenges. First of all, tracking task usually suffers from insufficient reliable positive instances, since the ground truth is only designated in the initial frame. Additionally, the online training is also computationally intensive which retards the speed. To customize CNN for visual tracking problem without compromising the performance, the reported methods tend to employ off-line trained networks for feature extraction. For example \cite{ma2015hierarchical}, the output of three layers \emph{conv3-4}, \emph{conv4-4} and \emph{conv5-4} from a pretrained VGG-Net-19 network are conserved and imported into a set of correlation filters for the encoding of target appearance. The scheme is based on the fact that early layers retain more fine-grained spatial details that can contribute to a precise localization and last layer usually encodes semantic abstraction that is robust to appearance variations.\\
\indent In practical implementation, it is almost impossible for the pretrained networks to process the whole image in each frame due to the huge computational load. Thanks to the essence of object tracking that the target's location in consecutive frames is highly correlated, most of the deep trackers only handle a cropped searching window centered at the location in previous frame. However, the window may fail to capture the object in the presence of abrupt or fast motion. To cope with this problem, we propose to introduce a shallow cue based prior map to guide the crop of the ROI window. Compared with deep features, although, shallow features have limited discriminative power, they cost less computational power to obtain and can competently act as an indicator to roughly describe the target. A top-down saliency structure using shallow cues has been verified to be feasible for visual tracking in \cite{li2016top}. In \cite{su2014abrupt}, the researchers propose to generate the saliency map by assigning weights to the three conspicuity maps in Itti and Koch's saliency model and conducting a weighted summation. Motivated by this work, we propose a novel prior map generation method by extending the three conspicuity maps to nineteen to enhance the descriptive capability. Moreover, the proposed method also considers the priori information in the previous frame as well as the constraints concerning target's size for the map building. Finally, the ROI is determined by performing template matching operation for all the candidate patches.\\
\indent In this paper, we utilize the pretrained VGG-Net-16 network for deep feature extraction as \cite{wang2015visual}. To avoid the distraction in final heat map, \cite{wang2015visual} proposes two networks using different VGG layers of output features to assist the target identification. But these two networks are both designed to learn the holistic appearance of the target which would be very susceptible to partial occlusion and distracters in some cases. On the other hand, part-based visual tracking scheme \cite{jia2012visual} has shown its great superiority in dealing with these kind of problems mentioned above. Therefore, we incorporate the part-based idea into deep tracking for the first time. Apart from the holistic learner network, four local networks are involved to learn one specific part of the target individually and generate four heat maps. The four local heat maps are then used to rectify the holistic map by removing the effect of distracters. In this manner, by adopting four additional hints to separate the real heat region from the fake ones, the risk of drifting is naturally reduced.\\
\indent To accommodate appearance variation during tracking, an online update procedure is commonly included in deep learning based methods. Target appearance in first frame \cite{wang2015visual} and current frame \cite{wang2015transferring} are usually collected as the positive samples. Although, the beginning frame contains reliable target depiction, it ignores the variation. And updating using current frame may cause overfitting occasionally to some detected false instances. To overcome this problem, we proposed a bio-inspired prioritised online update scheme. This scheme casts the selection of reliable positive instance as a label noise problem \cite{natarajan2013learning} by maximizing the joint probability which stands for an uncontaminated instance that is suitable for updating. The joint probability can be decomposed into two components, the selection possibility and a conditional probability which uses confidence value to evaluate the quality of the selected instance. The selection probability is designed by investigating both the temporal information and the sample quality in a created positive instance pool. A nonlinear weight allocation scheme is tailored to consider the temporal information inspired by bioinformatics that emphasizes the prediction results in beginning and recent frames and tend to forget the mid-term memories. In this manner, this method guarantees the selected positive instance is of high quality, at the same time, can reflect latest appearance of the target.\\
\indent The main contribution of this work is three-fold and can be summarized as: (i) A novel shallow cue guided global search algorithm is proposed to facilitate the determination of ROI window. The global search can provide a rough target location prediction for the deep trackers to release the computational burden without risking tracking failure when abrupt or fast motion happens. (ii) A mixed deep architecture including both holistic and local models is proposed. Apart from the traditional holistic network, four additional local networks are learnt from distinct components of the object. The holistic heat map is then rectified through the four local hints to make the proposed tracker more robust against distraction as well as overfitting. (iii) To guarantee the positive instance fed into the network for online update is uncontaminated and representative, we cast the selection of instance into a label noise problem. This design intends to emphasize samples from the beginning and current frames which reflects either reliable or up-to-date appearance information. The proposed update scheme ensures the network adaptive to the appearance variation of the target and avoid the performance degradation caused by inappropriate training samples.
\section{Prior Map Generation}
To achieve the tradeoff between computation load and the tracking accuracy, many CNN based approaches \cite{wang2015visual} \cite{wang2016stct} crop the ROI centered at the last target location before prorogating to the networks. This strategy inevitably causes the tracking failure when the target undergoes abrupt and fast motion and escapes from the ROI. To cope with this problem, this work adopts a prior map via shallow features to determine the ROI before deep feature extraction.
\subsection{Prior Map Building and Candidate ROI Generation}
Different from deep features, shallow features such as color, intensity, steerable pyramid subbands are more accessible and suitable for the rough localization of target. In \cite{judd2009learning}, the researcher investigated thirty-three features distributed from low-, mid- and high-level. The low-level features are proven to be able to depict the fundamental and general characteristics of the object and shows great efficiency compared with mid-, high-level features which need off-line training in advance. By considering the balance between efficacy and speed, nineteen low-level features are extracted to construct the final prior map. The first thirteen features employ the steerable pyramid subbands in four orientation and three scales and denoted as $F_{SPi}$, $i=1...13$. Moreover, four broadly tuned color channels ($F_R, F_G, F_B, F_Y$) as well as the intensity channel ($F_I$) are also taken into consideration. Finally, a channel of skin color $F_{SK}$ is involved since the tracker is very likely applied to track human targets. A feature map set is constructed by concatenating these nineteen features as $F_{FM}=[F_{SP1},...,F_{SP13},F_R, F_G, F_B, F_Y, F_I, F_{SK}]$. To further accelerate the algorithm, the arrived image are uniformly warped into $200\times200$ size first. Then we introduce a weight vector $w_{S}=[w_{SP1},...,w_{SP13},w_{R},w_{G},w_{B},w_{Y},w_{I},w_{Sk}]$ indicating the correlation degree between each feature map and the target area. The weights are determined in the first frame through an $L_2$ optimization problem as follow and kept fixed throughout the whole sequence.
\begin{eqnarray}
\label{wS}
\min_{w_{S}}\|X_{b}^{*}-F_{FM}^{'}w_{S}\|_{2}^2+\lambda_{S}\|w_{S}\|^2_{2},X_{b}^{*}=\begin{cases}1,(x,y) \in \phi_b\\
0,\text{otherwise} \end{cases}
\end{eqnarray}
where $F_{FM}^{'}$ is obtained by vectorizing the candidate feature set $F_{FM}$.  $\lambda_S$ is a penalty coefficient and $X_{b}^{*}$ represents a binary mask map where $\phi_b$ is a pixel set indicating the groundtruth bounding box in first frame. The optimal solution to (\ref{wS}) is computed as
\begin{eqnarray}
\label{solu}
w_{S}=(F^{'T}_{FM}F^{'}_{FM}+\lambda_{S}I)^{-1}F^{'T}_{FM}X_{b}^{*}
\end{eqnarray}
With the initialized weights $w_{S}$, a top-down saliency map $S^{'}_{map}$ is created through the weighted combination of the nineteen low-level feature maps as $S^{'}_{map}=\sum_{i=1}^{19} w_{Si}F_{FMi}$, where $F_{FMi}$ indicates the \emph{i}th element of $F_{FM}$. Different from the traditional center prior \cite{judd2009learning} which tend to believe human naturally arrange the object of interest near the center of the image, in this work, a revised center prior penalization is proposed to consider the spatial information in the previous frame into the prior map construction. To achieve this, we penalize the combined map $S^{'}_{map}$ with the distance to the center of target in the last frame as $S^{c}_{map}=C(p_c)\bigodot S^{'}_{map}$, where $\bigodot$ is the Hadamard product (element-wise product). $C(p_c)$ denotes a distance penalty matrix defined as $C_{ij}(p_c)=\delta_s\displaystyle\frac{Dis(p_{ij},p_c)}{\max{Dis(p_{ij},p_c)}}$, where $p_c$ and $p_{ij}$ represent the center of the estimated target in the last frame and a pixel position on $S^{'}_{map}$ respectively. $Dis(p_{ij},p_c)$ returns the Euclidean distance between $p_{ij}$ and $p_c$. $\delta_s$ is a tunable scaler and set as 2. In order to attenuate the noise and simplify the operation, the generated map $S^{c}_{map}$ is subject to binarization with a threshold $\sigma_b$ to produce the prior saliency map $S_{map}$. To proceed the analysis, we give the following observation.\\
\indent \emph{Observation 1}: The tracking target or part of it can map to a connected area on the saliency map $S_{map}$.\\
\indent The intuition behind this observation is that the whole body or parts of an object with highly discriminative features are usually contiguous. For instance, a face with skin color against the entire head with black hair and red mouth is salient which allows us to locate the rough position of a person's head by simply identifying connected areas with skin color on the saliency map. Under this assumption, an ``run-relabel" algorithm is employed to derive connected area set $\Omega_c=[a_{c1}, a_{c2},...,a_{cn}]$ with a predefined threshold $\sigma_s$ for size constraint, which indicates the minimum area that can be selected into set $\Omega_c$ and its value is proportional to the size of bounding box. Next, we return the geometric center of each area $a_{c1}$ to form $C_c=[c_{c1}, c_{c2},..., c_{cn}]$ and apply the bounding box with same scale and orientation in the last frame to every center element in $C_c$ to crop out the candidate particles $X_{c}=[x_{c1}, x_{c2},...,x_{cn}]$.\\
\indent To determine the final ROI region, we adopt a simple raw pixel template matching mechanism with the image patch $x^{*}_1$ extracted using the ground truth in the initial frame. The candidate with largest observation likelihood $c_i$ as calculated in (\ref{confroi}) is regarded as the predicted ROI in the current frame.
\begin{eqnarray}
\label{confroi}
c_i=\delta_c \exp(-\|x_{ci}-x^{*}_1\|_2^2)
\end{eqnarray}
where $\delta_c=0.01$ is used to ensure $c_i$ belonging to the range of [0,1]. In order to keep the algorithm as lite as possible, we fix the template $x^{*}_1$ to avoid improper update. This strategy may lead to drift problem if the target undergoes severe appearance variation and (\ref{confroi}) fails to give sufficient insight on the selection of correct ROI. To cope with this problem, we introduce an extra threshold $\sigma_c$ to judge whether the determined ROI is applicable. If the maximum confidence $c_{*}>\sigma_c$, the deep tracker is performed centered at the determined ROI, otherwise, the cropped image patch centered at the estimated location in previous frame is propagated into the deep tracker. %A summary on the determination of ROI is presented in Algorithm \ref{iteslv}.
%\begin{algorithm}
%  \caption{Prior Map Generation}
%  \label{iteslv}
%  \begin{algorithmic}[1]
%  \REQUIRE Input frame $y^{t}$, calculated feature weights $w_S$, location of target center in previous frame $p_c$, appearance template $x^{*}_1$.
%  \ENSURE Prior map $S_{map}$, determined ROI center $c_{ROI}$ for deep tracker.
%  \STATE Generate and concatenate nineteen raw feature maps to produce $F_{FM}$.
%  \STATE Predict the rough prior map $S_{map}^{'}$ with (\ref{salp}).
%  \STATE Modify the rough prior map by involving the temporal information of the target location in previous frame: $S^{c}_{map}=C(p_c)\bigodot S^{'}_{map}$ where $C(p_c)$ is subject to (\ref{tarpen}).
%  \STATE Binarize $S^{c}_{map}$ with threshold $\sigma_b$ to obtain $S_{map}$
%  \STATE Search the connected area on $S_{map}$ with predefined threshold $\sigma_s$ and extract the center location $C_c$ as well as the candidate patches $X_c$.
%  \STATE Determine the ROI by searching the maximum confidence $c_{*}$ of particles in $X_c$ using (\ref{confroi})
%  \IF {$c_{*}>\sigma_c$}
%  \STATE $c_{ROI}=c_{c*}$
%  \ELSE
%  \STATE $c_{ROI}=p_c$
%  \ENDIF
%  \end{algorithmic}
%\end{algorithm}
\section{Deep Tracking with Mixed Model}
Deep neural network such as CNN has shown its powerful capability in encoding the target appearance without human's guide. But, different from traditional classification and detection problems, the target object in visual tracking usually undergoes dynamic appearance variations due to partial occlusion, illumination changes and etc, which make it inaccurate sometimes impossible to reply on sole holistic information when developing trackers. In this regard, collaborative model \cite{zhong2014robust} with both local and holistic features have been proven to be competent in coping with the challenges mentioned above. Motivated by this, we propose a mixed model based deep tracking scheme to achieve a more robust performance.\\
\subsection{Proposed Network Architecture}
\begin{figure}[thpb]
  \centering
\subfigure {\includegraphics[width=0.8\hsize]{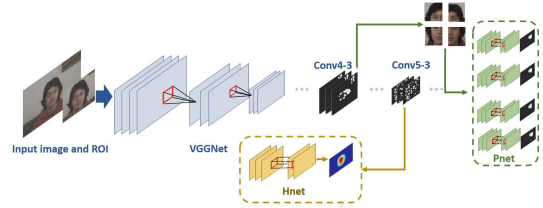}}
  \caption{The architecture of proposed mixed deep tracker} \label{supflow}
\end{figure}

CNN architectures such as AlexNet, VGG-Net, GoogLeNet are efficient in extracting discriminative and semantic features if they are trained with sufficiently large scale datasets. It is inappropriate to employ these deep networks directly on visual tracking task to estimate the target location since (i) we are lack of sufficient samples for training, (ii) the online training of such deep network is very time-consuming. To handle these contradictions, \cite{wang2015visual} proposes to use a 16-layer VGG-net for capturing the visual representation related to the object. Two small networks are trained specifically with holistic features on the top of VGG to produce the foreground heat map regression. In this paper, we adopt a similar framework as shown in Figure \ref{supflow}. \\
\indent The determined ROI of input frame is firstly propagated into a VGG-Net that has been pretrained on large-scale ImageNet dataset with category label \cite{deng2009imagenet}. With the forward propagation of the network, though, the semantical discrimination power increases, the spatial resolution reduces due to the pooling operation. For this reason, features obtained from Conv4-3 layer showing more intra-class discrimination power and higher resolution ($46\times46$) are more suitable for detail description, thus, extracted as the input of ``part-based" network \emph{Pnet}. \emph{Pnet} consists of four two-layer convolutional networks \emph{Pnet1}, \emph{Pnet2}, \emph{Pnet3} and \emph{Pnet4} using ReLU nonlinearity without any pooling operation. They are supposed to learn one specific part of the target and produce four corresponding heat maps $M_{P1}\sim M_{P4}$. On the contrary, features ($23\times23$) from Conv5-3 layer focusing on the semantic category information are fed into the holistic network \emph{Hnet}. \emph{Hnet} has the same architecture with \emph{Pnet}. It mainly handles the detection of the whole body of the object in target's category and generates a holistic heat map $M_{H}$. \\
\indent Before the initialization of \emph{Hnet} and \emph{Pnet}, we perform a feature map selection on the output of Conv4-3 and Conv5-3 in order to improve the detection accuracy by removing noisy features \cite{wang2015visual}. We similarly construct \emph{$Hnet_S$} and \emph{$Pnet_{S1}-Pnet_{S4}$} networks consisting of one dropout layer with dropout ratio of 0.3 and one convolutional layer with kernel size of $3\times3$ to predict the groundtruth heat map $M_{HT}$ and $M_{PT1}\sim M_{PT4}$. $M_{HT}$ is built by performing a 2-dimensional Gaussian distribution centered at the location of target object in the first frame. As for $M_{PT1}\sim M_{PT4}$, by considering the performance and time consumption simultaneously, we divide the target into four parts: top-left(TL), top-right(TR), bottom-left(BL) and bottom-right(BR) as shown in Figure \ref{supflow}. The intuition behind this is that we hope to train each $Pnet$ and $Pnet_{S}$ to learn one component of the object in order to handle the cases that some parts are subject to occlusion or severe deformation while others not. Each heat map $M_{PT}$ is constructed using one part through the 2-dimensional Gaussian distribution. The variance of the distribution is proportional to the corresponding object size. The loss functions are defined as $L_S=\|M-M_T\|^2$ where $M$ stands for the predicted maps, i.e. $M_{H}$, $M_{P}$. $M_T$ is the groundtruth maps. The impact of each input feature $f_i$ is evaluated through its effect on the loss function $L_S$. Quantitatively, a two-order Taylor expansion is employed to measure the impact as $\delta L_S=\sum_i \frac{\partial L_S}{\partial f_i} \delta f_i + \frac{1}{2} \sum_i \frac{\partial^2 L_S}{\partial f_i^2}(\delta f_i)^2 +\frac{1}{2} \sum_{i \neq j}\frac{\partial^2 L_S}{\partial f_i \partial f_j} \delta f_i \delta f_j$.
After approximating the Hessian matrix with a diagonal matrix and set $\delta f_i=0-f_i$, the impact of $f_i$ denoted as $\delta L_{Si}$ can be simplified as $\delta L_{Si}=-\frac{\partial L_S}{\partial f_i} f_i+\frac{1}{2}\frac{\partial^2 L_S}{\partial f_i^2} f_i^2$. The final score $SC_i$ is calculated by summing $\delta L_{Si}$ value at each pixel location $(x,y)$. The top $N_S=384$ features are selected as the salient ones. Readers are recommended to refer to \cite{wang2015visual} for more details about the feature selection strategy. With the salient features, the five networks (\emph{Hnet} and four \emph{Pnet}) are initialized with the same groundtruth maps respectively.\\
\subsection{Localizing the Target}
\label{S2b}
\begin{figure}[thpb]
  \centering
\subfigure {\includegraphics[width=1\hsize]{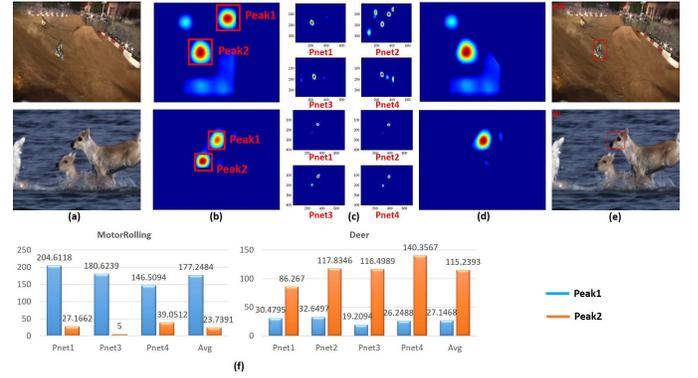}}
  \caption{Case study of proposed mixed deep tracking on \emph{MotorRolling} and \emph{Deer}. First and second row present: (a) input image, (b) heat map $M_H$ from Hnet, (c) heat maps $M_{P1}\scriptsize{\sim}M_{P4}$ from Pnet1$\tiny{\sim}4$, (d) rectified $M_H$, (e) tracking results. (f) voting distance between peaks in $M_p$s and $M_H$.} \label{casemixmodel}
\end{figure}

For most input images, the output holistic heat map $M_H$ is able to locate the target because of the strong discriminative power of deep features. However, $M_H$ may fail to provide a ``pure" indication with single peak on the heat map in the cases that distracters appear or insufficient online finetuning due to the lack of reliable samples as depicted in Figure \ref{casemixmodel} (b). For the \emph{MotorRolling} case, $M_H$ gives two peak areas peak1 and peak2. Peak2 corresponds to the real target, while peak1 incorrectly takes the grey wall as the target. This situation may be caused by the similar color and texture between the rider and the wall. In the case of \emph{Deer}, the holistic map provides two peak areas when the target approaches to another deer with similar appearance. Apparently, $M_H$ cannot be applied directly for target searching under this situation. Hence, when a new $M_H$ is generated, it is passed through a watershed approach \cite{beucher1992morphological} to search for regional maximum areas with peaks located at $(x_{hi}, y_{hi})$ where $i=1,2...n_h$. Further, we remove the areas which do not satisfy $M_H(x_{hi},y_{hi})\geq 0.8 \max (M_H(x_{hi},y_{hi}))$ and remains $n_h^{'}$ maximum areas. If $n_h^{'}>2$, it reveals that more than one peaks exist in the map $M_H$ and it should be rectified before localizing the target. In this paper, we propose to take the advantage of the four $M_P$s for the rectification since they can provide more intra-class specifics related with the target. Still, we obtain the peak locations of individual $M_P$ in the same way with $M_H$ and discard the maps with more than one peaks. The peaks' coordinates are recorded as $\{(x_{p1},y_{p2})...(x_{pn_p},y_{pn_p})\}$ where $n_p\leq4$. Next, the peaks from part-based maps vote for each peak in the holistic map with their Euclidean distance and the score $SH_i$ of each peak is calculated through the average of the distance. The voting procedure can be expressed as
\begin{eqnarray}
SH_i=\displaystyle\frac{\sum_{j=1}^{n_p}\sqrt{(x_{hi}-x_{pj})^2+(y_{hi}-y_{pj})^2}}{n_p},\ (i=1..n_h^{'})
\end{eqnarray}
The peak area with smallest score $SH_i$ is regarded as valid indication area and others are filled with zeros. Figure \ref{casemixmodel} present a clear example for the afore-mentioned rectification operation. The upper row image of (c) shows the heat maps using part-based hints. It can be observed that in $M_{p2}$, there is more than one peaks exist and we cannot tell which one corresponds to the real target, thus, this map is discarded. Moreover, though there also exist distracters in $M_{p4}$, the peak values of the distracters do not exceed $80\%$ of the maximum peak value, the peak in this map is valid for the voting stage. In this manner, the $M_H$ map can be efficiently rectified into a single-peak indicator as shown in (d).
\\
\indent We then perform particle filter on the rectified $M_H$ to decide the target location $\hat{X}(x,y,\sigma)$ which is described by the center coordinates as well as the scale. Given the center of the ROI $c_{ROI}$ and scale in the previous frame $\sigma^{t-1}$, the locations of target candidates are assumed to be subject to a Gaussian distribution centered at $X_{ROI}(c_{ROI},\sigma^{t-1})$ with a fixed diagonal covariance $\Psi$. The corresponding patches of candidates on the heat map are warped into the same size before the average heat value $v_i$ inside individual patch is calculated. The confidence of each candidate is finally derived by $C_i=v_i \sigma_i^{\gamma}$ where $\sigma_i$ stands for the scale of \emph{i}th candidate and $\gamma<1$ is an important coefficient that controls the scale compensation on $v_i$ in the confidence. It $\gamma$ is too large, obviously, candidates with large scale will be selected as the target. On the contrary, a small $\gamma$ will lead to a small scale of bounding box. Therefore, after trial and error, it is set as 0.7 in this paper. Candidate with highest confidence value $C_*$ is determined as the location of target. In practical implementation, we observe that although this scheme can produce an accurate center location, the bounding box is not tight. To remedy this, we make a minor modification on the scale $\sigma_*$ into $\sigma_{*}=\sigma^1_{*}(C_{*}/C_*^1)^{\lambda_{\sigma}}$ where $\sigma^1_{*}$ and $C_*^1$ are the scale and maximum confidence in the first frame respectively. $0\leq\lambda_{\sigma}\leq1$ adjusts the modification ratio. The efficacy of this modification is verified in the experiment section.
\section{Prioritised Online Update Scheme}
\begin{figure}[thpb]
  \centering
\subfigure {\includegraphics[width=0.95\hsize]{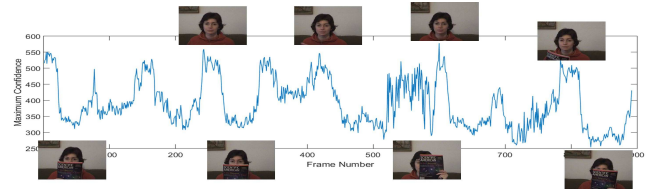}}
  \caption{Case study on video sequence of \emph{FaceOcc1} containing 892 frames for quantitative measurement of $P(\xi=1|t)$ using $C_*$. When the wave troughs appear, the face is usually partially or fully occluded as the lower image row shows. Rising of $C_*$ is caused by recovery from the occlusion. Crest occurs when the face is completely unoccluded. } \label{mcof}
\end{figure}

In order to adapt to the appearance variation of the target, the top added networks should be online updated. Due to the possible in-plane rotation of the target, it is difficult to determine the relative position and orientation of the four parts of \emph{Pnet}s in current frame. To avoid an overcomplicated update, we adopt a conservative update strategy for the \emph{Pnet}s by fixing them after the initialization. The update scheme in this section focuses on the holistic network \emph{Hnet}. \\
\indent In this paper, the positive training instances are selected among the tracking results in individual frame. The key condition to ensure an efficient update is to search ``good" training samples which are not contaminated by occlusions or misalignment. We cast this problem into a label noise problem \cite{natarajan2013learning} by introducing a joint probability $P(y^{*}_t,\xi=1)$. $y^{*}_t$ stands for the estimated result in frame $t$. $\xi=1$ denotes the sample is not contaminated and ready for update, otherwise $\xi=0$. $P(y^{*}_t,\xi=1)$ describes the likelihood that the selected positive sample within the tracked frames is suitable for the online updating. Considering the chain rule, we have
\begin{align}
P(y^{*}_t,\xi=1)&=P(t,\xi=1)=P(\xi=1|t)P(y^{*}_t)
\end{align}
Our goal is to appropriately design the conditional probability $P(\xi=1|t)$ and selection possibility $P(y^{*}_t)$ such that $P(y^{*}_t,\xi=1)$ is as large as possible. Firstly, the challenge is how to estimate $P(\xi=1|t)$, in other words, we hope to quantitatively measure the quality of a sample. In this regard, the optimal confidence value $C_*$ is employed to achieve the measurement based on the observation that a lower $C_*$ is usually caused by (i) full or partial occlusion (ii) severe scale variation. A case study on \emph{FaceOcc1} is conducted as shown in Figure \ref{mcof} to explain the observation (i). When no occlusion occurs, the optimal confidence usually corresponds to a high value, otherwise, a lower value is rewarded. For (ii), the reason is straightforward that the shrink of scale will definitely lead to a vanish of the value as well as the size of corresponding heat area on the heat map. Small scale of the target inevitably contains less information and is considered not ``good" enough for updating. With the analysis above, a sample is determined as contaminated when either (i) or (ii) situation occurs. Therefore, we predict $P(\xi=1|t)$ as $P(\xi=1|t) \propto C_*^t$. To ensure a high confidence value of each selected $y_t^*$, we create a pool $\mathbb{Y}^*=\{y_{t(1)}^*, y_{t(2)}^*, ..., y_{t(N_y)}^*\}$ containing $N_y$ samples with large $C_*$. The insertion of new element $y_t^*$ should satisfy either (i) $C_*^{t}>\min \{C_*^{t(1,..,N_y)}\}$ or (ii) $C_*^{t}/\max \{C_*^{t(1,..,N_y)}\}>0.85$. Once the condition is satisfied, we replace $y^*_{t(\arg \min_i \{C_*^{t(1,.i.,N_y)}\})}$ with $y^*_{t}$.\\
\indent As for the selection probability $P(y^{*}_t)$, we design it by mainly considering two criteria: (i) uncontaminated samples possess larger selection probability, (ii) good image temporal location is rewarded with larger selection probability. (i) focuses on the evaluation of one specific sample and can be similarly quantified using $C_*$. (ii) bases on the intuition that samples extracted from the beginning few frames contain the most accurate and reliable message on the object, moreover, the recent few images ahead of the current frame contain the most up-to-date appearance description about the object. Thus, we propose the following quadratic-like weight allocation function for the purpose of emphasizing the most original and recent samples and assign smaller weights for the mid-sequence ones.
\begin{eqnarray}
%\label{weiup}
\label{weiup} W_i^{t}=\displaystyle\frac{4(1-\theta)}{t(t-2)}\mathscr{T}^2(i)+\displaystyle\frac{4(t+1)(\theta-1)}{t(t-2)}\mathscr{T}(i)+\displaystyle\frac{t-4\theta+2}{t-2}
\end{eqnarray}
where $W_i^t$ denotes the temporal weight for \emph{i}th element among $\mathbb{Y}^*$ in frame \emph{t}. $1\leq \mathscr{T}(i)\leq t$ records the frame number when the \emph{i}th sample is added into $\mathbb{Y}^*$. $\theta<1$ adjusts the smallest value of the weights which always occur at the middle of the tracked video sequence. Recalling the specific sample quality measurement index $C_*^{t(1,..,N_y)}$, a comprehensive evaluation index $I^t_{i}$ is derived by $ I^t_i=W_i^{t}*C_*^{t(i)}/\max \{C_*^{t(1,..,N_y)}\}$. In this manner, we rationally assign the weight for each sample in the pool considering specific (individual quality) and general (sequential characteristics) measurements. The selection probability $P(y_t^*)$ is finally determined based on $I^t_i$, i.e. $P(y_t^*)=\displaystyle\frac{I^t_i}{\sum_{i=1}^{N_{y}}I^t_i}$. The design of $P(\xi=1|t)$ and $P(y_t^*)$ is able to ensure a high quality positive instance for the update. Different from positive sample generation, negative sample is captured directly by removing the estimated target in current frame. Since the distracters in background may belong to distinct categories and is subject to high uncertainty, there is no need to also create a negative sample pool like $\mathbb{Y}^*$. Using the negative samples in current frame is sufficient and easy-operating.\\
\indent \textbf{Update condition:} we investigate two issues to determine if the update should be executed, which are the tracking quality of the target and the existence of distracters in background. Concretely, update is activated upon the satisfaction of the following two conditions simultaneously.\\
\indent (i) Select one positive sample $y^*_{t(n)}$ in $\mathbb{Y}^*$ randomly according to the possibility $P(y_t^*)$ and $C_*^{t(n)}>2C_*^{t}$.\\
\indent (ii) There appears more than one peaks in heat map $M_{H}$.\\
\indent In order to avoid updating the network over-frequently, we check these two conditions every ten frames. Once the update is required, the network is online tuned by minimizing the following cost function.
\begin{align}
\label{cofun}
\notag &\min  \beta_W \|W_H\|_{F}^2+\sum_{x,y}\big[\mathscr{L}^{n}(x,y)Tru\big(M_H^{n}(x,y)-M^{n}_T\\
&(x,y)\big)^2+\big(1-\mathscr{L}^{t}(x,y)\big)Tru\big(M_H^{t}(x,y)-M_T^{t}(x,y)\big)^2\big]
\end{align}
where $W_H$ denotes the convolutional weights in Hnet. $\mathscr{L}^{i}(x,y)$ stands for the label of pixel $(x,y)$ in \emph{i}th frame (1 for foreground, 0 for background). $M_T^{i}$ is the Gaussian distributed groundtruth map. $ Tru(\bullet)$ represents the truncated loss to accelerate the online updating process \cite{li2016deeptrack}. It is defined as
\begin{eqnarray}
\label{trunfun}
Tru(\bullet)=|\bullet|\bigg(1-\textbf{l}\bigg[|\bullet|\leq \displaystyle\frac{\epsilon}{(k+\mu\phi^{i}(x,y))}\bigg]\bigg)
\end{eqnarray}
This truncation is based on the observation that the tracking performance is more sensitive to the prediction error on positive samples than negative ones.
\section{Experiments}
The proposed SDT is implemented in MATLAB and run at 1.5 fps on an Intel Core i7-4710HQ 2.5GH PC with 16GB memory and NVIDIA Geforce GTX 860M GPU. The deep networks are built and trained on a wrapper of Caffe framework. The thresholds for prior map generation are set to $\sigma_b=0.2$, $\sigma_{s}=0.4*w*h$ and $\sigma_c=0.2$, where $w$ and $h$ are the width and height of current bounding box. The kernel sizes of the two convolutional layers in $Hnet$ are $9\times9$ and $5\times5$ with respective padding of 4 and 2. The four $Pnet$s share the same architecture with $Hnet$. The output features from Conv5-3 are resized into $46\times46$ by linear interpolation before importing to $Hnet$. $Hnet$ and $Pnet$s are initialized in the first frame using back-propagation for 100 iterations. The number of particles sampled for target localization is 700. The parameter $\theta$ in (\ref{weiup}) is 0.7 and the truncation parameters in (\ref{trunfun}) are set to $\epsilon=e_{\text{max}}$, $k=20$, $\mu=30$. The capacity of positive sample pool $\mathbb{Y}^*$ is $N_y=10$.\\
\indent Since the generation of prior map depends on all the three channels of a image, we conduct the test on twenty challenging color video sequences in \cite{wu2015object}. And the results are compared with fifteen state-of-the-art trackers including L1APG \cite{bao2012real}, IVT \cite{ross2008incremental}, Frag \cite{adam2006robust}, KCF \cite{henriques2015high}, Struck \cite{hare2016struck}, MTT \cite{zhang2012robust}, ASLA \cite{jia2012visual}, KMS \cite{comaniciu2003kernel}, CXT \cite{dinh2011context}, CSK \cite{henriques2012exploiting}, DFT \cite{sevilla2012distribution}, LOT \cite{oron2015locally}, SCM \cite{zhong2014robust}, TLD \cite{kalal2012tracking} and FCNT \cite{wang2015visual} for qualitative and quantitative study.
\subsection{Ablation Study}
\begin{figure}[thpb]
  \centering
\subfigure {\includegraphics[width=1\hsize]{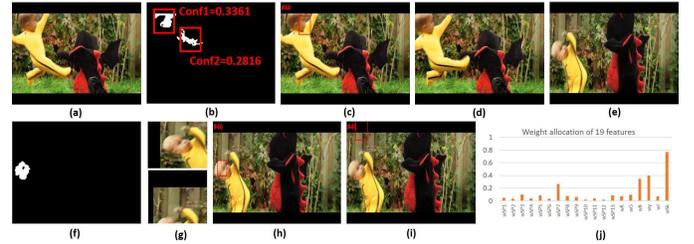}}
  \caption{Efficacy validation of prior map guidance on video sequence of \emph{DragonBaby}. (a) Input image of frame 44, (b) $S_{map}$ map for frame 44, (c) tracking result in frame 44, (d) input image of frame 45, (e) input image of frame 46, (f) $S_{map}$ map for frame 46, (g) determined ROI of using prior map (upper figure) and not (lower figure), (h) tracking result in frame 46 using prior map, (i) tracking result in frame 46 without prior map, (j) weight allocation of nineteen features in prior map generation.} \label{draabl}
\end{figure}

This section conducts several tests on the main components of SDT for the performance verification. Figure \ref{draabl} presents the experiment on the video sequence of \emph{DragonBaby}. (a) and (b) are the input image of frame 44 and the produced saliency map $S_{map}$. Since we are supposed to track the head of the baby in this task, the weight of the channel of skin color $w_{Sk}$ dominates over other channels according to the histogram in (j). Moreover, due to the similar color essence, weight of yellow channel $w_Y$ also possesses relatively large value (half of $w_{Sk}$). This leads to the existence of distracter in (b) because of the yellow leaves in background. Thanks to the template matching operation, which allows us to separate these two candidate ROIs with different confidence $c_*$. The upper salient area which represents the real target has higher confidence of 0.3361, while, the fake one only possesses 0.2816. In this manner, the ROI can be accurately captured in the presence of distraction and excellent tracking performance is guaranteed in (c). (d) and (e) show two consecutive frames 45 and 46 where the target is subject to abrupt motion from the upper left corner to the middle. (f) reveals that the location of target after the large motion can be uniquely indicated in the prior map $S_{map}$ and no template matching separation is required. (g) shows the determined ROI with and without prior map's guide. The lower ROI is cropped centered at the location in frame 45 and it can be observed that target disappears in the patch, thus, the tracking fails as shown in (i). On the contrary, $S_{map}$ can help to relocate the center of ROI as demonstrated in upper image of (g) and in turn contributes to an accurate tracking in (h).\\
\indent To investigate the necessity and performance of the prioritised update scheme for $Hnet$, a series of experiments on different video sequences using distinct update schemes are carried out and the results are shown in Table \ref{stusms}. ``NoUdt", ``Udt1f" and ``UdtCf" stand for ``no update", ``update using positive sample in first frame" and ``update using positive sample in current frame". The negative samples in (\ref{cofun}) among all the experiments are extracted from the current frame as in the proposed update scheme. The performance is evaluated via overlap rate and center error and the best performance is highlighted in red font. Generally, the proposed scheme can achieve better accuracy compared with other three schemes. Specifically, for the cases that the target undergoes severe appearance variation such as \emph{Dog}, \emph{MotorRolling} and \emph{Skiing}, SDT shows its superiority since it can rationally select the positive samples without involving too much noise and ensure the network adaptive to the changes. For those no severe appearance variation occurs such as \emph{Couple} and \emph{Deer}, ``NoUdt" can already handle the tracking well. The involvement of online update will not degrade the performance using the proposed scheme. When the target is subject to occlusions as in \emph{Bird2}, the two insertion conditions for positive sample pool $\mathbb{Y}^*$ guarantee that only high-quality samples are available as the candidates for update, hence, the network's degradation is alleviated.
\begin{table}[thpb]
\centering
%\tiny
\scriptsize
\caption{Study of different update schemes}\label{stusms}
\begin{tabular}{c c c c c c c c c c c c c c c c c} % centered columns (4 columns)
    \hline\hline
       & \multicolumn{2}{c}{NoUdt} & \multicolumn{2}{c}{Udt1f} & \multicolumn{2}{c}{UdtCf} & \multicolumn{2}{c}{SDT}\\ \hline\hline
       &OR & CE&OR & CE&OR & CE&OR & CE\\ \hline
       Bird2&0.68&11.67&0.69&10.98&0.71&10.26&\color{red}{0.72}&\color{red}{10.23}\\ \hline
       Couple&\color{red}{0.69}&5.08&0.66&5.02&0.65&4.72&\color{red}{0.69}&\color{red}{4.27} \\ \hline
      Deer&\color{red}{0.72}&7.78&\color{red}{0.72}&7.70&\color{red}{0.72}&7.63&\color{red}{0.72}&\color{red}{7.58}\\ \hline
      Dog&0.49&12.07&0.54&7.56&0.46&\color{red}{6.83}&\color{red}{0.55}&6.89\\ \hline
      MotorR & 0.58&14.89&0.61&14.44&0.59&15.16&\color{red}{0.65}&\color{red}{14.13}\\ \hline
      Skiing & 0.51&5.33&0.51&4.22&\color{red}{0.55}&\color{red}{3.64}&\color{red}{0.55}&3.99\\ \hline
      Average & 0.61&9.47&0.62&8.32&0.61&8.04& \color{red}{0.65}& \color{red}{7.85}\\
      \hline
     \multicolumn{9}{l}{\tiny Note: OR and CE are short for overlap rate and center error respectively.}
    \end{tabular}
\end{table}
\subsection{Qualitative Evaluation}
\begin{figure*}[thpb]
  \centering
\subfigure [] {\includegraphics[width=0.9\hsize]{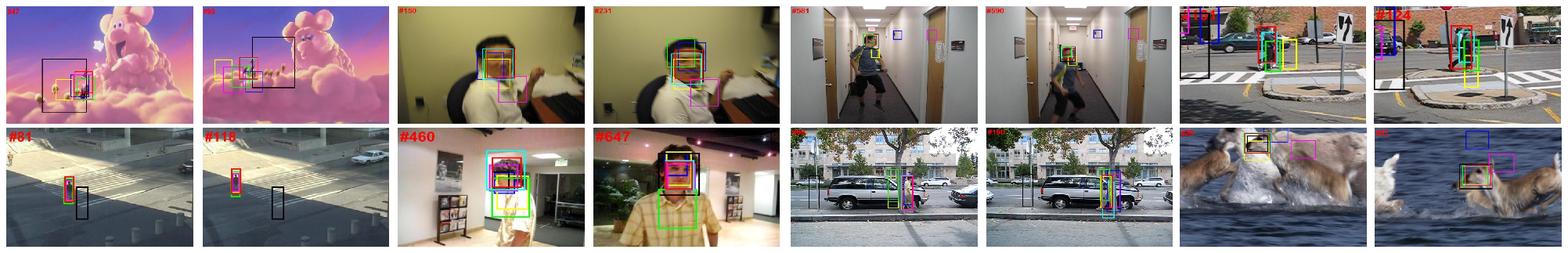} \label{1qua} }
\subfigure [] {\includegraphics[width=0.9\hsize]{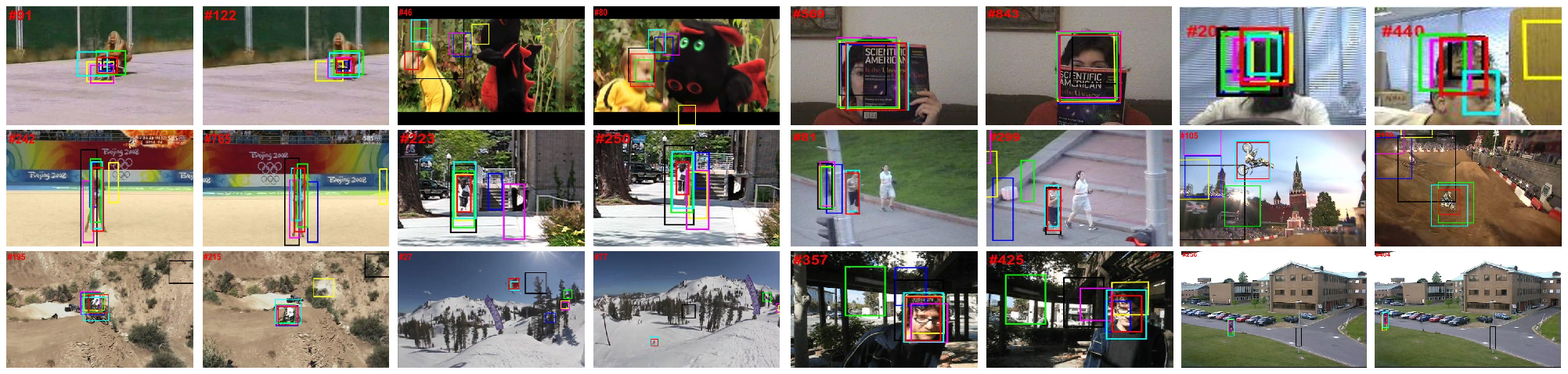} \label{2qua} }
\subfigure {\includegraphics[width=0.6\hsize]{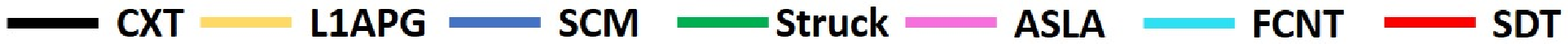}}
  \caption{Tracking result screenshots of seven trackers. (a) Video sequences of \emph{Bird2}, \emph{BlurFace}, \emph{Boy}, \emph{Couple}, \emph{Crossing}, \emph{David}, \emph{David3}, \emph{Deer}. (b) Video sequences of \emph{Dog}, \emph{DragonBaby}, \emph{FaceOcc1}, \emph{Girl}, \emph{Gym}, \emph{Human7}, \emph{Jogging}, \emph{MotorRolling}, \emph{MountainBike}, \emph{Skiing}, \emph{Trellis}, \emph{Walking}.} \label{quastudy}
\end{figure*}

Qualitative investigation is carried out on twenty challenging video sequences with six state-of-the-art trackers. The result is reported in Figure \ref{quastudy} and the detailed analysis follows below.\\
\indent \textbf{Background clutter}: cluttered background brings the challenge by degrading the contrast between foreground and background. In \emph{Couple}, the running cars, jeep as well as the tree in the background may lead to drifting of the trackers. Only the proposed SDT is able to perform favorably by the end of the sequence since deep features equip the tracker with excellent discriminative power to separate the background and foreground. In \emph{Deer}, the sequence undergoes both distracters and fast motions. Algorithms based on raw pixel or PCA template such as L1APG, SCM and ASLA are not efficient enough to handle the similar context objects nearby. All the tracking by detection methods (Struck, FCNT, SDT) can complete the task perfectly due to the discriminative capability of nonlinear classifier. Moreover, the four part-based heat maps facilitate to locate the real target from the distracters as mentioned in \ref{S2b}. In \emph{MountainBike}, the biker rides across the gap with varying postures. The rocks and bushes inside the gap as interference factors bring extra difficulty to the trackers. CXT and L1APG drifts after the biker's landing. FCNT does not produce a tight bounding box in the last few frames since it include noise in the background into the heat map due to imperfect online update.\\
\indent \textbf{Abrupt motion and motion blur}: To tackle fast motion is usually very challenging for visual tracking task since trackers tend to search the target near the location in previous frame to achieve a good tradeoff between performance and efficiency. As can be seen in the first figure of \emph{DragonBaby}, all the trackers fail to follow the target due to the abrupt motion except for our method with prior map's guide. An acceptable result can still be obtained when the target is subject to sudden scale variation as the second image shows. This benefits from the minor modification of scale parameter $\sigma_*$ after the determination of target's location. Moreover, the tracking task becomes more complicated since abrupt or fast motion will always lead to motion blur which makes it ineffective to perform template matching for traditional methods. The robustness of the deep tracking against blur image can be demonstrated by \emph{BlurFace}, \emph{Boy} and \emph{Human7}. The target can be captured throughout the whole sequence although slight out-of-plane rotation and scale variation occurs in \emph{Boy} and \emph{Human7} at the same time.\\
\indent \textbf{Full or partial occlusion}: unlike some particle filter based methods such as L1APG using trivial templates to reconstruct the occluders, tracking-by-detection methods have no specific designed mechanism to cope with occlusions. The tracker may not be able to produce a reliable holistic heat map when occlusion occurs. Therefore, the proposed mixed model shows its superiority in dealing with partial occlusion since it can learn the unoccluded parts to rectify the holistic map for better performance. For example the last frame in \emph{Girl}, when the girl's face is partially blocked by the man, the proposed tracker can still identify the real target through the assistance of part-based heat maps. While, the holistic feature based FCNT method drifts to the man's face incorrectly. Other part-based algorithm such as ASLA also performs well in this case. When full occlusion happens, we cannot indicate the target on both the image and the heat maps. Under this situation, to avoid drifting we involve a little trick by fixing the bounding box when extreme low confidence $C_*^t$ is detected. Then, the tracker will recapture the target after the full occlusion disappears with increased $C_*^t$. This trick is sample but efficient as shown in the second image of \emph{David3}. FCNT has drifted to the lower half of the man after fully blocked by the tree, while, the proposed SDT can still achieve the tracking. Finally, the prior map and deep features' discriminative power makes it possible to track the object again in case of drifting due to the occlusion as shown in \emph{David3} and \emph{Jogging}.\\
\indent \textbf{Non-rigid object deformation}: non-rigid deformation indicates the severe appearance variation when human, animal moves. Trackers need to develop proper update scheme to capture the latest changes to improve the adaptiveness. In \emph{Dog}, although most of the algorithms can track the target when the dog runs towards the woman, some of them may drift a lot when the dog shaking its tail under the coverage of her shadow. The slight illumination change brings fatal interference to these trackers. In \emph{Gym}, the athlete performs varying gymnastic movement in the field throughout the video, the proposed tracker can always follow the torso of hers and some trackers may slightly miss the target at the end of the task. In \emph{Skiing}, non-rigid deformation happens together with scale variation and background clutter when the skier flights over the hillock in front of the camera. Only the two deep learning based methods SDT and FCNT can finish the tracking. Since we perform scale modification after the target localization, a tighter bounding box can be achieve as shown in the last figure.\\
\indent \textbf{Illumination variation and rotation}: The proposed tracker also shows pretty good capability in handling illumination changes as reported in \emph{Trellis} when the guy walks under the light varying condition. CXT which depends on the context elements exploration is very susceptible to the changes and drifts almost at the beginning of the test. Other methods employing raw pixels or haar-like feature are all sensitive to the illumination variation as the result report. The ability that the proposed scheme can handle in-plane or out-of-plane rotation attributes to the architecture. The holistic map learns robust high-level semantic information and the part-based learners are in charge with components of the target which is especially significant in the presence of in-plane rotation as demonstrated in \emph{MotorRolling}. Moreover, SDT can also handle mirror rotation in \emph{Brid2}, \emph{David3} or even more comprehensive rotation scenario in \emph{David}, \emph{Girl}.
\subsection{Quantitative Comparison}
In order to achieve a comprehensive evaluation, quantitative comparison experiment is conducted with other fifteen state-of-the-art algorithms in terms of success score and precision score \cite{wu2015object}. The experiment results are reported in Table \ref{succterr} and \ref{cnterr} where the top three scores are highlighted in red, blue and green fonts.\\
\begin{table*}[thpb]
\centering
\tiny
%\scriptsize
%\footnotesize
\caption{Average success scores}\label{succterr}
\begin{tabular}{c c c c c c c c c c c c c c c c c} % centered columns (4 columns)
    \hline\hline
       & SDT&$L_1$APG & IVT &Frag&KCF&Struck&MTT&ASLA&KMS&CXT&CSK&DFT&LOT&SCM&TLD&FCNT  \\ \hline\hline
       Bird2&\textcolor{red}{0.879}&0.101&0.434&0.232&0.252&0.494&0.090&\textcolor{blue}{0.798}&0.262&0.101&0.454&0.697&0.070&0.676&0.303&\textcolor{green}{0.777} \\ \hline
       BlurF&\textcolor{green}{0.990}&0.385&0.119&0.661&0.140&0.612&0.357&0.091&0.780&\textcolor{red}{1}&\textcolor{blue}{0.998}&0.281&0.359&0.135&\textcolor{red}{1}&\textcolor{blue}{0.998}\\ \hline
       Boy&\textcolor{blue}{0.962}&0.566&0.294&0.518&0.318&\textcolor{green}{0.938}&0.435&0.420&0.794&\textcolor{blue}{0.962}&0.814&0.481&0.629&0.438&0.601&\textcolor{red}{0.968}\\ \hline
        Couple& \textcolor{red}{0.779}&\textcolor{green}{0.550}&0.085&0.492&0.235&0.514&0.492&0.085&0.064&0.500&0.085&	 0.085&	0.450&	0.078&	 0.178&	\textcolor{blue}{0.671}\\
       \hline
       Cross&\textcolor{green}{0.817}&	\textcolor{red}{1}&	0.191&	0.366&	0.300&	0.800&	0.216&	 \textcolor{blue}{0.991}&	0.233&	 0.316&	 0.166&	0.508&	0.450&	\textcolor{blue}{0.991}&	 0.425&	 0.816\\ \hline
       David&0.540&	0.428&	\textcolor{green}{0.581}&	0.057&	0.199&	0.189&	0.233&	 \textcolor{red}{0.646}&	 0.023&	0.482&	0.515&	0.189&	 0.084&	0.318&	 0.518&	 \textcolor{blue}{0.634}\\ \hline
        David3&\textcolor{green}{0.889}&	0.329&	0.507&	0.678&	\textcolor{red}{0.956}&	0.337&	 0.095&	 \textcolor{blue}{0.934}&	 0.718&	 0.111&	0.189&	0.662&	0.722&	 0.456&	0.107&	0.865\\ \hline
        Deer&\textcolor{blue}{0.901}&	0.760&	0.028&	0.126&	\textcolor{green}{0.774}&	 \textcolor{red}{0.957}&	0.704&	 0.042&	 0.309&	0.478&	 \textcolor{red}{0.957}&	0.309&	0.042&	 0.042&	 0.281&	\textcolor{blue}{0.901}\\ \hline
        Dog&\textcolor{green}{0.307}&	0.078&	0.094&	0.039&	0.047&	0.070&	0.063&	 0.189&	 0.047&	0.401&	 0.047&	0.047&	 \textcolor{blue}{0.362}&	\textcolor{red}{0.378}&	 0.244&	0.039\\
        \hline
        DragB&\textcolor{red}{0.788}&	0.238&	0.230&	0.362&	0.053&	0.088&	 0.132&	0.132&	 0.398&	0.336&	 0.212&	0.11&	 \textcolor{green}{0.495}&	 0.097&	0.070&	\textcolor{blue}{0.725}\\
        \hline
        Occ1&0.924&	\textcolor{green}{0.992}&	0.871&	\textcolor{red}{1}&	0.956&	0.970&	0.698&	 0.915&	 0.873&	 0.651&	\textcolor{red}{1}&	 0.698&	0.245&	\textcolor{blue}{0.998}&	0.191&	 0.937\\
        \hline
        Girl&\textcolor{blue}{0.816}&	0.440&	0.168&	0.488&	0.482&	0.302&	\textcolor{red}{0.858}&	 \textcolor{green}{0.606}&	 0.246&	0.598&	 0.294&	0.184&	0.470&	0.324&	 0.258&	0.328\\ \hline
        Gym&\textcolor{blue}{0.144}&	0.003&	0.003&	0.136&	\textcolor{red}{0.160}&	0.015&	0.010&	 0.006&	 0.109&	0.088&	0.010&	0.014&	 0.011&	0.087&	 \textcolor{green}{0.138}&	0.109\\ \hline
        Human7&\textcolor{blue}{0.692}&	\textcolor{green}{0.516}&	0.264&	0.128&	0.420&	0.152&	 0.156&	 0.240&	 0.092&	0.276&	0.160&	0.152&	 0.404&	 0.240&	\textcolor{red}{0.820}&	0.144\\ \hline
        Jogging&\textcolor{green}{0.883}&	0.198&	0.221&	0.517&	0.224&	0.195&	 0.218&	0.224&	 0.169&	 \textcolor{blue}{0.951}&	0.221&	 0.215&	0.087&	 0.172&	\textcolor{red}{0.954}&	0.853\\ \hline
        MotorR&\textcolor{red}{0.640}&	0.030&	0.042&	0.073&	0.054&	\textcolor{green}{0.134}&	 0.048&	 0.067&	 0.054&	0.018&	0.048&	0.048&	 0.030&	 0.042&	0.115&	\textcolor{blue}{0.481}\\ \hline
        MoutB&\textcolor{blue}{0.930}&	0.723&	0.877&	0.122&	0.232&	0.693&	 0.653&	0.833&	 0.434&	0.276&	 \textcolor{green}{0.921}&	0.350&	 0.622&	 0.473&	0.263&	\textcolor{red}{0.982}\\ \hline
        Skiing&\textcolor{red}{0.432}&	0.086&	0.074&	0.037&	0.049&	0.037&	 \textcolor{green}{0.098}&	 \textcolor{blue}{0.111}&	 0.012&	 \textcolor{blue}{0.111}&	0.049&	0.049&	0.012&	 0.049&	0.061&	 \textcolor{red}{0.432}\\ \hline
        Trellis& \textcolor{red}{0.874}&	0.551&	0.253&	0.233&	0.163&	\textcolor{green}{0.692}&	 0.145&	 0.688&	 0.203&	0.479&	0.209&	 0.479&	0.261&	 \textcolor{blue}{0.739}&	0.411&	0.643\\ \hline
        Walking&0.719&	0.534&	\textcolor{blue}{0.885}&	0.371&	0.368&	0.446&	 0.640&\textcolor{red}{	 0.995}&	 0.284&	0.216&	0.393&	0.410&	 \textcolor{green}{0.851}&	 0.830&	0.293&	0.405\\ \hline\hline
        Average &\textcolor{red}{0.745}&	0.425&	0.311&	0.332&	 0.319&	 0.431&	0.317&	 \textcolor{green}{0.451}&	0.305&	 0.418&	0.387&	 0.299&	0.333&	0.378&	0.362&	 \textcolor{blue}{0.635}\\
        \hline
    \end{tabular}
\end{table*}
\begin{table*}[thpb]
\centering
\tiny
%\footnotesize
%\scriptsize
\caption{Average precision scores}\label{cnterr}
\begin{tabular}{c c c c c c c c c c c c c c c c c} % centered columns (4 columns)
    \hline\hline
       & SDT&$L_1$APG & IVT &Frag&KCF&Struck&MTT&ASLA&KMS&CXT&CSK&DFT&LOT&SCM&TLD&FCNT  \\ \hline\hline
       Bird2& \textcolor{red}{0.990}&0.131&	0.484&	0.313&	0.343&	0.545&	0.090&	 \textcolor{green}{0.878}&	 0.393&	0.282&	0.515&	0.717&	 0.080&	0.808&	 0.313&	\textcolor{blue}{0.949}\\ \hline
       BlurF&0.988&	0.355&	0.109&	0.630&	0.123&	0.612&	0.316&	 0.091&	 0.754&	\textcolor{red}{1}&	 \textcolor{green}{0.995}&	0.281&	0.389&	 0.137&	\textcolor{blue}{0.998}&	 \textcolor{green}{0.995}\\ \hline
       Boy&\textcolor{red}{1}&	0.604&	0.332&	0.574&	0.382&	\textcolor{red}{1}&	0.445&	0.440&	 0.990&	 \textcolor{red}{1}&	 0.843&	0.485&	 0.666&	0.440&	\textcolor{red}{1}&	\textcolor{red}{1}\\ \hline
        Couple& \textcolor{red}{1}&	0.585&	0.085&	\textcolor{blue}{0.907}&	0.257&	 \textcolor{green}{0.757}&	 0.642&	0.107&	 0.107&	0.578&	 0.085&	0.085&	0.635&	 0.085&	0.221&	 \textcolor{red}{1}\\
       \hline
       Cross&0.993&	\textcolor{red}{1}&	0.983&	0.400&	\textcolor{red}{1}&	\textcolor{red}{1}&	0.250&	 \textcolor{red}{1}&	\textcolor{red}{1}&	 0.575&	\textcolor{red}{1}&	 0.683&	0.635&	\textcolor{red}{1}&	 0.583&	\textcolor{red}{1} \\ \hline
       David& 0.889&	0.796&	\textcolor{red}{1}&	0.110&	0.787&	0.322&	0.343&	 \textcolor{red}{1}&	 0.522&	 \textcolor{red}{1}&	0.498&	 0.354&	0.284&	0.422&	\textcolor{blue}{0.955}&	 \textcolor{green}{0.921}\\ \hline
        David3& \textcolor{green}{0.976}&	0.345&	0.754&	0.789&	\textcolor{red}{1}&	0.337&	0.107&	 0.674&	 \textcolor{green}{0.976}&	0.158&	 0.658&	0.746&	\textcolor{blue}{0.988}&	0.674&	 0.115&	0.936
\\ \hline
        Deer&\textcolor{blue}{0.972}&	0.816&	0.028&	0.154&	0.816&	\textcolor{red}{1}&	0.760&	 0.042&	 0.535&	 \textcolor{green}{0.831}&	 \textcolor{red}{1}&	0.309&	0.183&	0.084&	0.309&	 \textcolor{red}{1}\\ \hline
        Dog&\textcolor{red}{1}&	0.881&	0.165&	0.858&	0.992&	0.976&	\textcolor{red}{1}&	0.968&	 0.622&	 \textcolor{red}{1}&	 \textcolor{red}{1}&	0.724&	0.897&	\textcolor{red}{1}&	0.622&	0.984\\
        \hline
        DragB&\textcolor{red}{0.921}&	0.256&	0.327&	0.477&	0.070&	0.194&	 0.168&	0.283&	 0.486&	0.566&	 0.212&	0.123&	 \textcolor{green}{0.681}&	 0.168&	0.088&	\textcolor{blue}{0.867}\\
        \hline
        Occ1&0.618&	\textcolor{green}{0.746}&	0.649&	\textcolor{red}{0.980}&	0.681&	0.566&	0.318&	 0.657&	 0.543&	0.389&	 \textcolor{blue}{0.947}&	0.622&	0.253&	\textcolor{blue}{0.947}&	 0.109&	 0.626\\
        \hline
        Girl& \textcolor{red}{1}&	0.596&	0.452&	0.652&	0.864&	\textcolor{blue}{0.974}&	 \textcolor{red}{1}&	\textcolor{red}{1}&	 0.536&	 \textcolor{green}{0.936}&	0.552&	0.296&	0.640&	 0.356&	0.90&	 0.914\\ \hline
        Gym& 0.930&	0.020&	0.509&	0.942&	0.794&	0.651&	0.255&	 0.573&	 \textcolor{red}{0.967}&	0.749&	 0.520&	0.233&	 \textcolor{green}{0.958}&	0.645&	 0.799&	\textcolor{blue}{0.959}\\ \hline
        Human7& 0.896&	0.948&	0.288&	0.484&	0.472&	\textcolor{red}{1}&	\textcolor{red}{1}&	 0.240&	0.572&	 \textcolor{red}{1}&	0.656&	 0.172&	0.460&	0.284&	\textcolor{red}{1}&	 0.808
\\ \hline
        Jogging&\textcolor{red}{0.978}&	0.228&	0.224&	0.661&	0.234&	0.231&	 0.228&	0.231&	 0.224&	 \textcolor{green}{0.960}&0.228&	0.215&	 0.599&	 0.228&	\textcolor{blue}{0.973}&	 \textcolor{blue}{0.973}\\ \hline
        MotorR&\textcolor{blue}{0.829}&	0.030&	0.036&	0.073&	0.048&	\textcolor{green}{0.365}&	 0.048&	 0.042&	 0.054&	0.024&	0.042&	0.042&	 0.048&	 0.042&	0.103&	\textcolor{red}{0.865}\\ \hline
        MoutB& \textcolor{blue}{0.987}&	0.864&	\textcolor{red}{1}&	0.140&	0.478&	0.964&	\textcolor{red}{1}&	 0.907&	0.666&	 0.280&	 \textcolor{red}{1}&	0.350&	0.693&	\textcolor{green}{0.982}&	0.280&	 \textcolor{red}{1}\\ \hline
        Skiing& \textcolor{red}{1}&	\textcolor{green}{0.135}&	0.111&	0.030&	0.074&	0.037&	0.123&	 \textcolor{green}{0.135}&	 0.111&	0.209&	 0.098&	0.074&	0.024&	0.074&	 0.061&	 \textcolor{red}{1}\\ \hline
        Trellis& \textcolor{blue}{0.974}&	0.783&	0.253&	0.374&	0.353&	0.789&	 0.230&	0.718&	 0.353&	 0.790&	0.810&	0.506&	0.309&	 \textcolor{green}{0.906}&	0.479&	\textcolor{red}{0.985}\\ \hline
        Walking&\textcolor{red}{1}&	\textcolor{red}{1}&	\textcolor{red}{1}&	0.987&	\textcolor{red}{1}&	 \textcolor{red}{1}&	\textcolor{red}{1}&	 \textcolor{red}{1}&	0.995&	0.235&	 \textcolor{red}{1}&	 \textcolor{red}{1}&	\textcolor{red}{1}	 &\textcolor{red}{1}	&0.915&	 \textcolor{red}{1}
\\ \hline\hline
        Average&\textcolor{red}{0.947}&	0.556&	0.439&	0.527&	 0.538&	 \textcolor{green}{0.666}&	0.466&	 0.549&	0.570&	 0.628&	0.633&	 0.401&	 0.521&	0.514&	0.541&	 \textcolor{blue}{0.939}
\\
        \hline
    \end{tabular}
\end{table*}
\indent It can be observed that the proposed tracker performs favorably on most of the video clips and outperforms other state-of-the-art methods. The superior result benefits a lot from the robust attributes of deep features. The same situation holds for FCNT. Furthermore, the tracking framework introduced in this paper guarantees a performance improvement compared with FCNT especially in terms of success scores. Apart from the deep learning based methods, ASLA possesses the third place in average success scores test, which demonstrates that part-based model is always efficient to facilitate the algorithms in dealing with challenging tracking tasks. It is worth noticing that we test all the video clips with three channels since the prior map relies on color images. However, if there is priori knowledge that the target will not undergoes severe abrupt motion or other challenges that need to determine the ROI in advance such as pedestrian tracking task or certain security surveillance systems, the proposed system can also be deployed on greyscale images without the prior map generation phase. \\
%\begin{figure*}[thpb]
%  % Requires \usepackage{graphicx}
%  \centering
%\subfigure {\includegraphics[width=1\hsize]{SuccRate.eps}}
%  \caption{Success and precision score plots of OPE. Seven trackers: CXT, L1APG, SCM, Struck, ASLA, FCNT and SDT are evaluated in terms of 10 attributes. IV: illumination variation, SV: scale variation, OCC: occlusion, DEF: deformation, MB: motion blur, FM: fast motion, IPR: in-plane rotation, OPR: out-of-plane rotation, BC: background clutter, LR: low resolution.} \label{attriope}
%\end{figure*}
\section{Conclusion}
In this paper, a shallow feature guided deep tracking algorithm has been developed with mixed models. To dynamically determine the ROI fed into the deep networks, nineteen hand-crafted shallow features are employed with learnt weights to generate a prior map. The pre-determined ROI helps to handle target's abrupt and fast motion. It is then passed into a novel mixed model based deep tracker with holistic learner to detect the semantic information in the image and part-based learners to handle the low-level discriminative features. The part-based maps facilitate to rectify the holistic heat map in the presence of occlusions, distracters and other interference factors. Finally, a prioritised update scheme is introduced for the online finetuning to alleviate degradation of the networks. A series of comprehensive experiments are conducted and demonstrate the superiority of the proposed tracker.

\scriptsize
\bibliographystyle{ieeetr}
\bibliography{ref}
\end{document}